\documentclass[runningheads]{llncs}

\usepackage{amsmath,amssymb,amsfonts}

\usepackage{etoolbox}
\usepackage{caption}
\usepackage[section]{placeins}
\usepackage{float}

\usepackage{multirow}

\usepackage{textcomp}
\usepackage{bbding}
\usepackage{pifont}

\usepackage[normalem]{ulem}

\usepackage{graphicx}    
\usepackage{xcolor}      
\usepackage{algorithmic}

\usepackage{cite}
\usepackage{orcidlink}

\usepackage{hyperref}

\hypersetup{
    colorlinks=true,      
    linkcolor=black,      
    citecolor=black       
}

\begin{document}
\title{CMFDNet: Cross-Mamba and Feature Discovery Network for Polyp Segmentation}
\titlerunning{Polyp Segmentation with Cross-Mamba Decoder}
\author{Feng Jiang\textsuperscript{1}\orcidlink{0009-0000-2878-7827}, 
Zongfei Zhang\textsuperscript{2}\orcidlink{0009-0000-6702-8807}, Xin Xu\textsuperscript{3}}
\authorrunning{F. Jiang et al.}
\institute{
\textsuperscript{1}School of Computer Science and Technology, Changsha University of Science and Technology, Changsha 410114 China\\
\textsuperscript{2}SCOT – Optimal Sourcing Systems Amazon.com Services LLC Seattle,\\ Washington USA\\
\textsuperscript{3}School of Mechanical and Electrical Engineering, China University of Mining and Technology, Beijing 100083  China\\
\email{24208051816@stu.csust.edu.cn}
}
\maketitle             
\begin{abstract}
Automated colonic polyp segmentation is crucial for assisting doctors in screening of precancerous polyps and diagnosis of colorectal neoplasms. Although existing methods have achieved promising results, polyp segmentation remains hindered by the following limitations, including: (1) significant variation in polyp shapes and sizes, (2) indistinct boundaries between polyps and adjacent tissues, and (3) small-sized polyps are easily overlooked during the segmentation process. Driven by these practical difficulties, an innovative architecture, CMFDNet, is proposed with the CMD module, MSA module, and FD module. The CMD module, serving as an innovative decoder, introduces a cross-scanning method to reduce blurry boundaries. The MSA module adopts a multi-branch parallel structure to enhance the recognition ability for polyps with diverse geometries and scale distributions. The FD module establishes dependencies among all decoder features to alleviate the under-detection of polyps with small-scale features. Experimental results show that CMFDNet outperforms six SOTA methods used for comparison, especially on ETIS and ColonDB datasets, where mDice scores exceed the best SOTA method by 1.83\% and 1.55\%, respectively.

\end{abstract}

\section{Introduction}
Among all malignant tumors worldwide, colorectal cancer (CRC) exhibits both high lethality and considerable incidence~\cite{b17}. Colonic polyps are considered precursor lesions of colorectal cancer, making early detection and removal critically important~\cite{b43}. Colonoscopy can provide information on the accurate location and appearance of polyps, thereby effectively preventing colorectal cancer~\cite{b32}. However, due to the significant variability of colonic polyps in colonoscopy, traditional manual inspection methods are highly dependent on the operator’s subjective experience, which may lead to a certain degree of missed diagnoses and misdiagnoses. Therefore, leveraging computer vision for the automatic segmentation of colonic polyps has emerged as a prominent direction in the field of medical image analysis.

\begin{figure}[!t]
\centering
\includegraphics[width=1\textwidth]{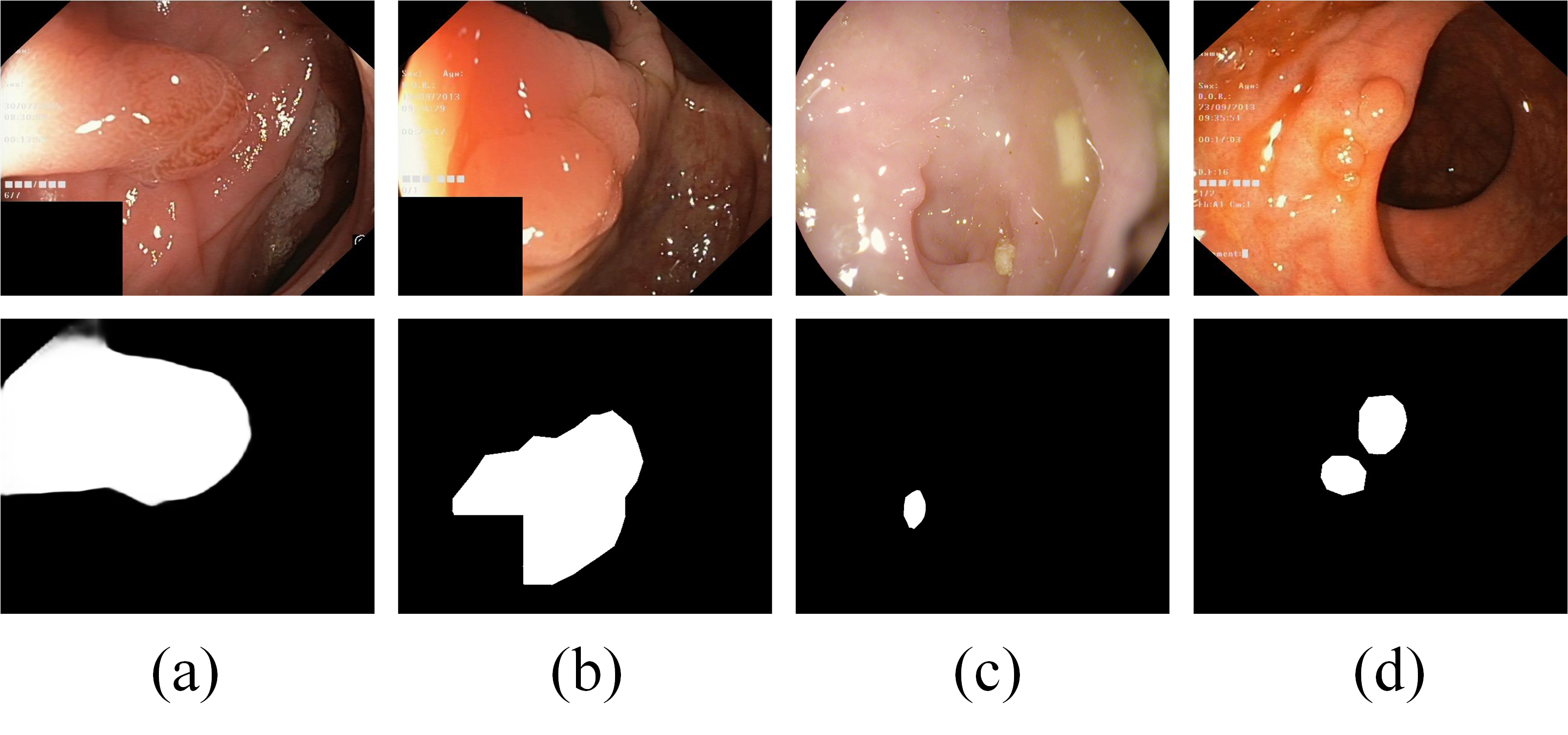}
\caption{Representative visual samples of polyps.}
\label{example}
\end{figure}

To design an effective colonic polyp segmentation method, it is essential to first analyse the characteristics of colonic polyp images and identify the challenges in segmentation. As shown in Fig.~\ref{example}, Fig.~\ref{example}(a)(b)(c) exhibit significant variations in polyp sizes and shapes. The polyp shown in Fig.~\ref{example}(c) is of a small size, making it susceptible to omission during the segmentation process. Fig.~\ref{example}(d) contains two separate polyps with blurred boundaries between them and the surrounding environment.

To address the three challenges mentioned above, a Cross-Mamba and Feature Discovery Network for Polyp Segmentation is proposed. The CMD module, based on Mamba, fuses deep and shallow features through cross-scanning, progressively recovering fine-grained features and reducing blurry boundaries. The MSA module employs three parallel branches, each utilizing convolution kernels of different sizes to improve the ability to identify polyps exhibiting diverse morphological and dimensional characteristics. The FD module establishes dependencies among all decoder features, thereby reducing the loss of small polyps. This study introduces the following key innovations: (1) A newly designed segmentation framework, CMFDNet, is specifically tailored for colonic polyp segmentation; (2) Three key modules—CMD, MSA, and FD—are designed to address three major challenges in polyp segmentation. The CMD module, based on Mamba, introduces a cross-scanning approach to fuse deep and shallow features, thereby reducing blurry boundaries. The MSA module adopts a multi-branch parallel structure to strengthen the segmentation performance for polyps with heterogeneous scales and contours. The FD module integrates features from all decoders to minimize missed detections of polyps at small scales; (3) Compared to six representative state-of-the-art approaches, the proposed network demonstrates superior performance, achieving mDice scores that surpass the best SOTA method by 1.83\% on ETIS and 1.55\% on ColonDB, respectively.

\section{Related Work}
\subsection{Medical Image Segmentation}
Medical image segmentation has significantly advanced with deep learning, particularly through CNNs, Transformers, and State Space Models. Early approaches such as Fully Convolutional Networks (FCN)~\cite{b20} introduced end-to-end pixel-wise prediction, laying the foundation for deep learning-based segmentation. U-Net~\cite{b21} further improved segmentation accuracy by introducing a dual-stage structure composed of encoding and decoding modules connected via intermediate shortcuts. With the emergence of Transformer, Vision Transformer (ViT)~\cite{b34} marked the initial application of the Transformer paradigm to visual recognition tasks, enhancing accuracy for high-resolution medical image segmentation. Recently, State Space Models have emerged as a possible alternative to Transformer architectures due to their linear complexity. Vision Mamba~\cite{b35} first introduced Selective State Space Models (SSMs) into image recognition. VMamba~\cite{b36} introduced the VSS block with a 2D Selective Scan (SS2D) module for improved contextual modeling and feature extraction. 

\subsection{Polyp Image Segmentation}
Colon polyp segmentation is a critical branch of medical image segmentation, facing challenges such as marked heterogeneity in polyp morphology and dimensional scale, blurred boundaries, and the loss of small polyps. Driven by these limitations, numerous enhanced strategies have been introduced in the literature. To accommodate the pronounced heterogeneity in polyp morphology and dimensional scale, Polyp-Mamba~\cite{b36} employs a Scale-Aware Semantics (SAS) module to enhance multi-scale feature learning for robust segmentation. LHONet~\cite{b16} integrates a Rough Outline Generation (ROG) unit, which extracts approximate structural cues to facilitate the initial detection of polyps with diverse appearances. For boundary blurriness between polyps and surrounding tissues, PraNet~\cite{b11} integrates a Reverse Attention (RA) mechanism that progressively sharpens boundary representations through iterative refinement. BDG-Net~\cite{b38} addresses boundary ambiguity by generating a boundary distribution map via the Boundary Distribution Generation Module (BDGM) and using it within the decoder to guide accurate boundary localization throughout the segmentation process. To address the challenge of small-sized polyps being easily missed, To mitigate the risk of missing small polyps, LSSNet~\cite{b27} employs a Multiscale Feature Extraction (MFE) block that captures high-resolution local patterns from shallow layers, enabling the retention of fine lesion details during progressive downsampling in the encoder. MoE-Polyp~\cite{b45} enhances small polyp detection through a mixture-of-experts decoder that adaptively integrates boundary-, spatial-, and global-aware expert features to alleviate information loss during decoding. 

\section{Method}
\begin{figure}[!t]
\centering
\includegraphics[width=1\textwidth]{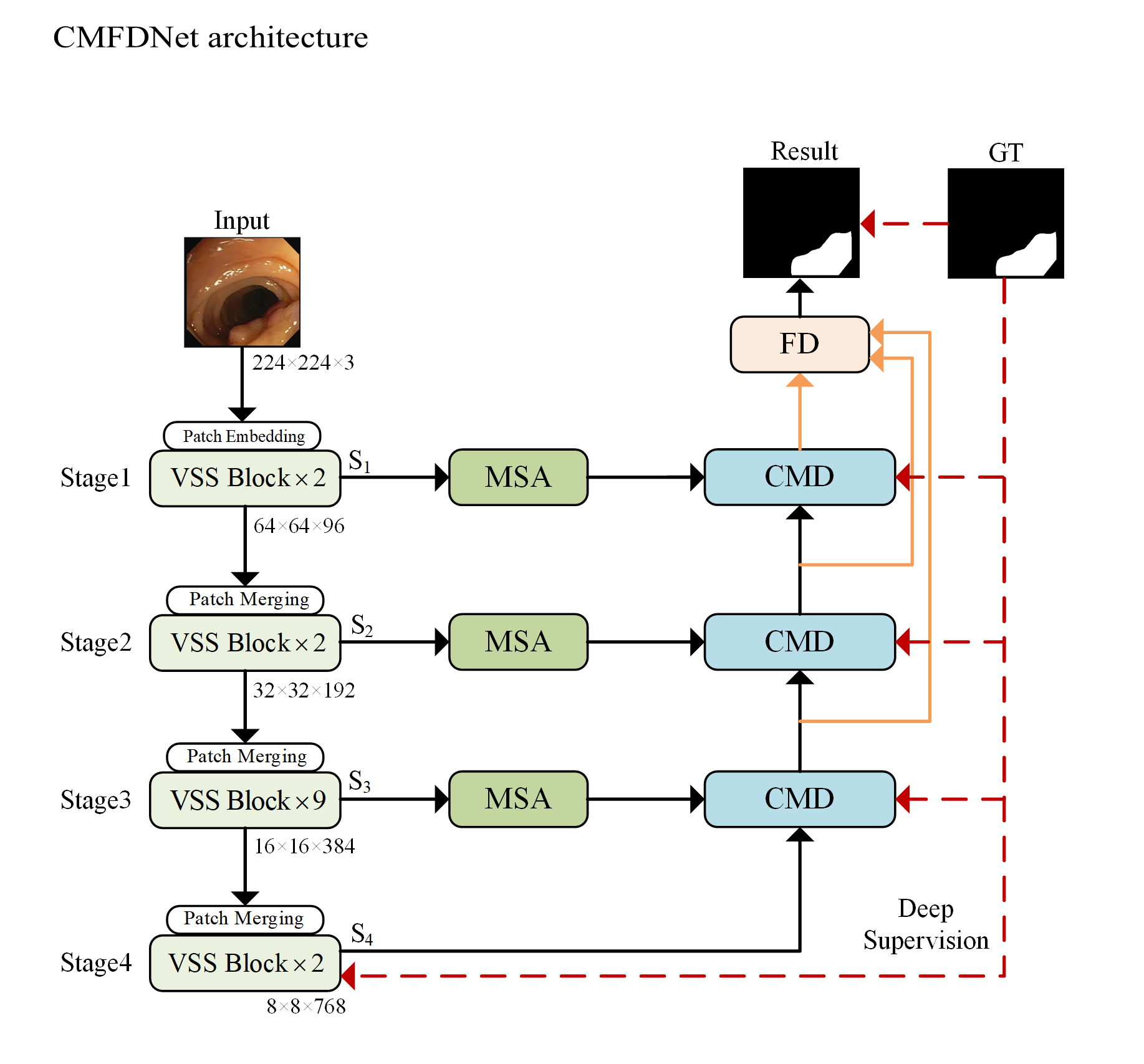}
\caption{The architecture of CMFDNet.}
\label{cmfdnet}
\end{figure}

The architecture of CMFDNet is shown in Fig.~\ref{cmfdnet}. VMamba-Tiny~\cite{b36}, pre-trained in advance, is employed as the core feature extractor within the encoder stage. The CMD module, as the decoder, employs a cross-scanning approach to fuse deep and shallow features, progressively recovering fine-grained features and enhancing the recognition capability for blurry boundaries. The MSA module adopts an inverted bottleneck structure~\cite{b1}, with three parallel branches performing multi-scale feature extraction to enhance the recognition ability for polyps of various sizes and shapes, while also functioning as an intermediate conduit that links feature representations from the encoder to the decoder stage. The FD module fuses all features from the CMD module to to alleviate the omission of diminutive polyps during the segmentation process. During training, deep supervision is employed, which is not used during testing.

\subsection{Cross-Mamba Decoder Module}
\begin{figure}[!t]
\centering
\includegraphics[scale=1.15]{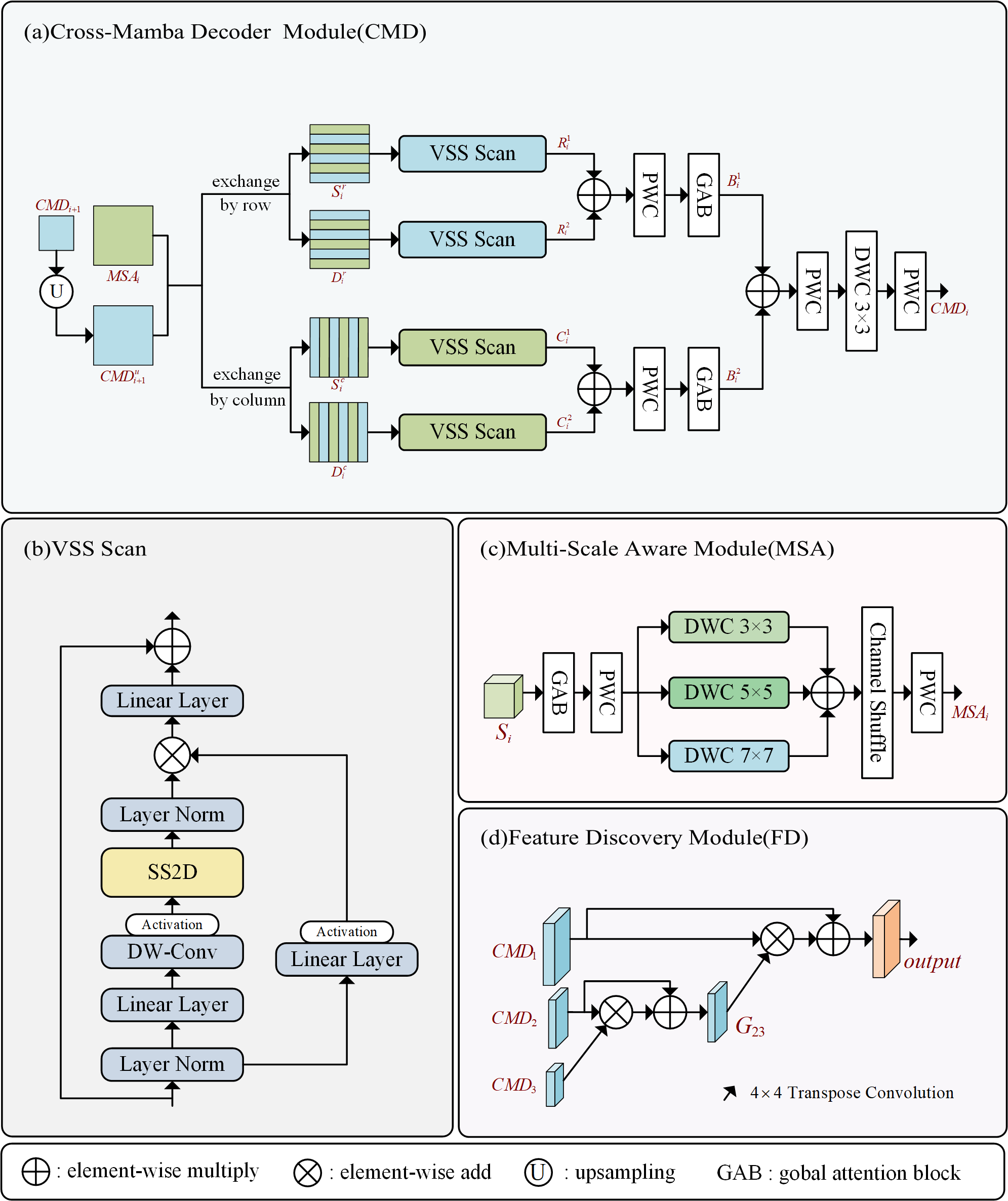}
\caption{(a) CMD architecture. (b) VSS Scan architecture. (c) MSA architecture. (d) FD architecture.}
\label{mynet}
\end{figure}

To mitigate boundary ambiguity in polyp regions, the Cross-Mamba Decoder (CMD) module is proposed, as shown in Fig.~\ref{mynet}(a). The CMD module performs cross-scanning on features from the MSA module and deeper CMD module, gradually fusing deep semantic features from the deeper CMD module with shallow local features from the MSA module, thereby progressively recovering fine-grained features.

\begin{figure}[!h]
\centering
\includegraphics[width=0.9\textwidth]{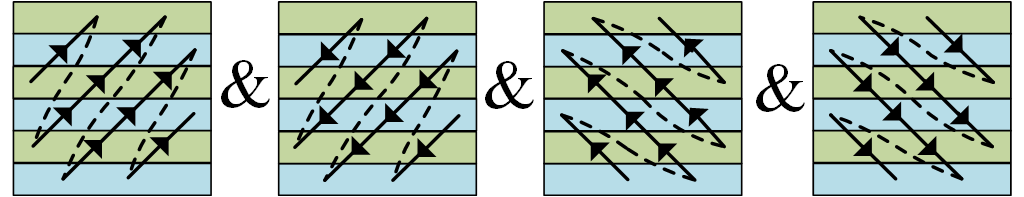}
\caption{Four types of diagonal scanning methods in SS2D}
\label{scanway}
\end{figure}
First, the features $CMD_{i+1}$ from the deeper CMD module are upsampled to obtain $CMD_{i+1}^{u}$. Then, a row-wise pixel exchange operation is performed on the features $CMD_{i+1}^{u}$ and $MSA_{i}$ from the i-th MSA module to obtain $S_{i}^{r}$ and $D_{i}^{r}$. Similarly, a column-wise pixel exchange operation is applied to obtain $S_{i}^{c}$ and $D_{i}^{c}$. The exchanged features $S_{i}^{r}$, $D_{i}^{r}$, $S_{i}^{c}$, and $D_{i}^{c}$ are processed by the VSS Scan using cross-scanning to obtain the fused features $R_{i}^{1}$, $R_{i}^{2}$, $C_{i}^{1}$, and $C_{i}^{2}$, respectively. As shown in Fig.~\ref{mynet}(b), the VSS Scan is similar to the VSS Block~\cite{b36}. However, the scanning methods adopted in its SS2D component differ from those of the VSS Block. The SS2D in the VSS Scan employs four types of diagonal scanning~\cite{b46} to achieve cross-scanning. The specific scanning methods are illustrated in Fig.~\ref{scanway}. This process can be expressed by the following formulas:
\begin{gather}
S_i^r, D_i^r = \mathit{Row\_Exchange}(MSA_i, CMD_{i+1}^{u}) \tag{1} \\
S_i^c, D_i^c = \mathit{Column\_Exchange}(MSA_i, CMD_{i+1}^{u}) \tag{2} \\
R_{i}^{1} = \mathit{VSS\ Scan}(S_{i}^{r}); R_{i}^{2} = \mathit{VSS\ Scan}(D_{i}^{r}) \tag{3} \\
C_{i}^{1} = \mathit{VSS\ Scan}(S_{i}^{c});C_{i}^{2} = \mathit{VSS\ Scan}(D_{i}^{c}) \tag{4} 
\end{gather}
Subsequently, $R_{i}^{1}$ and $R_{i}^{2}$ are added element-wise, followed by pointwise convolution to fuse the channel features. Then, feature enhancement is performed using the gobal attention block, resulting in $B_{i}^{1}$. The same operation is applied to $C_{i}^{1}$ and $C_{i}^{2}$ to obtain $B_{i}^{2}$. Finally, $B_{i}^{1}$ and $B_{i}^{2}$ are added element-wise and further feature extraction is performed through a series of convolution operations. This process can be expressed by the following formulas:
\begin{gather}
B_{i}^{1} = GAB(PWC(R_{i}^{1}+R_{i}^{2})) \tag{5} \\
B_{i}^{2} = GAB(PWC(C_{i}^{1}+C_{i}^{2})) \tag{6} \\
CMD_{i} = PWC(DWC_{3\times3}(PWC(B_{i}^{1}+B_{i}^{2}))) \tag{7}
\end{gather}
where $i=1,2,3$. $PWC$ is the pointwise convolution, $GAB$ means the gobal attention block, ${DWC}_{k\times k}$ refers to applying depthwise convolution using a k × k-sized kernel. 
\begin{figure}[!h]
\centering
\includegraphics[width=1\textwidth]{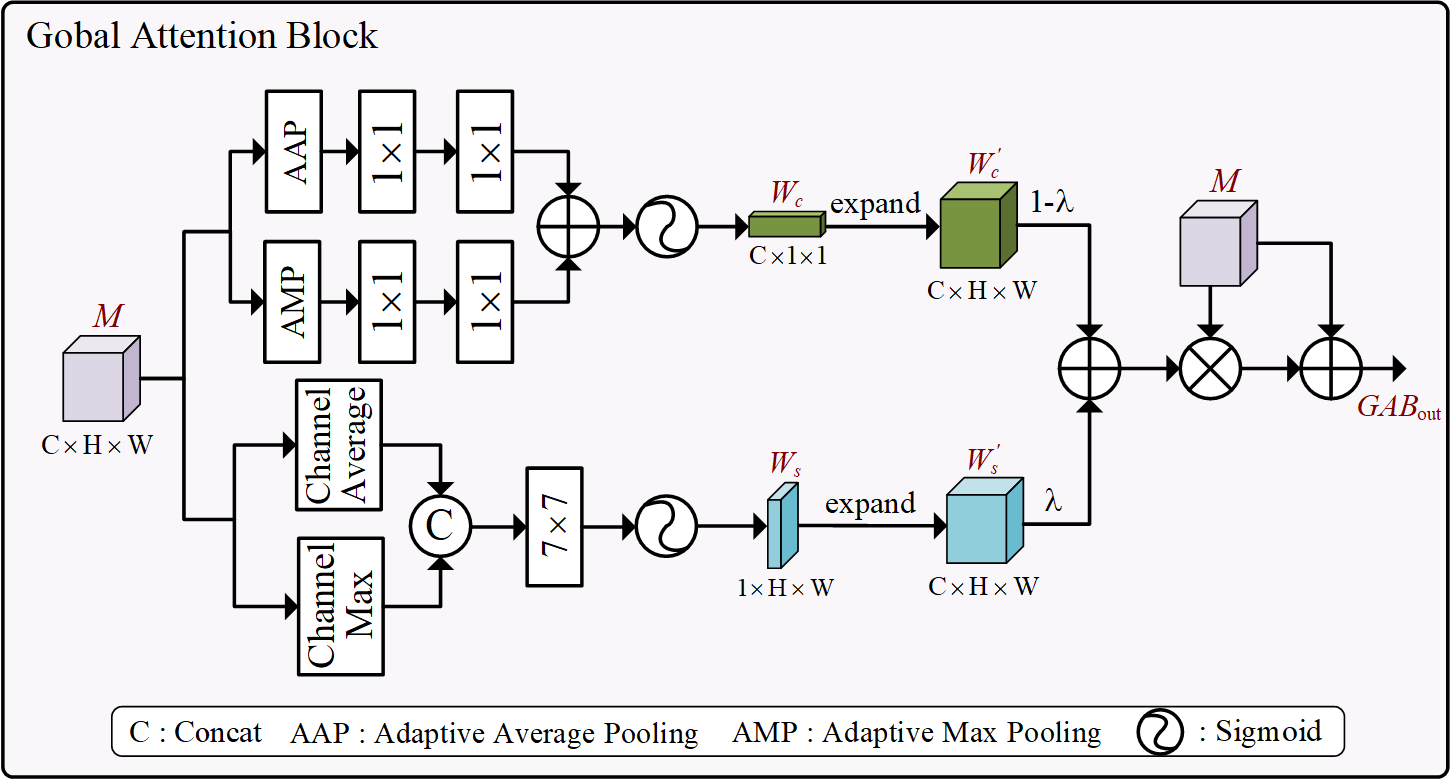}
\caption{The architecture of GAB.}
\label{gobal}
\end{figure}

\subsubsection{Gobal Attention Block.}
In serially connected channel and spatial attention modules, such as CBAM~\cite{b48}, the latter module may inadvertently diminish or override the features emphasized by the former, potentially leading to an amplification of noise, particularly when the noise distribution is nonuniform. To address this issue, the Global Attention Block (GAB) is proposed, as depicted in Fig.~\ref{gobal}. GAB adopts a parallel channel and spatial attention structure, which enables a more comprehensive integration of both types of feature information and flexibly suppresses noise effects through weighted fusion. For the input features $M$, the upper and lower branches utilize channel and spatial attention blocks~\cite{b49}, respectively, to obtain weights \( W_c \) and \( W_s \), which are then expanded to match the dimensions of \( M \), resulting in \( W_c' \) and \( W_s' \). Subsequently, a parameter $\lambda$ is introduced to adaptively weight $W_{c}^{'}$ and $W_{s}^{'}$ to obtain $W_{cs}$, representing the relative importance of channel and spatial information. Finally, $W_{cs}$ is multiplied with the input features $M$, followed by a skip connection, to produce the output $GAB_{out}$. This process can be expressed as follows:
\begin{gather}
W_{cs} = (1-\lambda) \times W_{c}^{'} + \lambda \times W_{s}^{'} \tag{8}\\
GAB_{out} = W_{cs}\times M+M \tag{9} 
\end{gather}
Where $\lambda$ $\in$ (0,1) is obtained by applying a sigmoid function to a learnable parameter, and is initialized to 0.5.
\subsection{Multi-Scale Aware Module}
To enhance the recognition capability for polyps of different sizes and shapes, the Multi-Scale Aware (MSA) module is proposed, which adopts an inverted bottleneck structure with three parallel branches for multi-scale feature extraction. As shown in Fig.~\ref{mynet}(c), the input features $S_i$ from the i-th stage first undergo feature enhancement through the gobal attention block, followed by a pointwise convolution that doubles the channel dimension, resulting in features $O_{i}$. Then, a three-branch parallel structure is applied, where each branch uses depthwise convolution with different kernel size: 3 × 3, 5 × 5, 7 × 7, resulting in $O_{i}^1$, $O_{i}^2$, and $O_{i}^3$. Subsequently, an element-wise addition is performed to aggregate the outputs of the three parallel branches. As depthwise convolution overlooks the relationships among channels, a channel shuffle~\cite{b47} operation is used  to incorporate relationships among channels. Afterward, an extra pointwise convolution is introduced to halve the channel dimension, which also encodes dependency among channels. This process can be expressed as follows:
\begin{gather}
O_{i} = PWC(GAB(S_{i})) \tag{10} \\
O_{i}^{1} = DWC_{3\times3}(O_{i}) \tag{11} \\
O_{i}^{2} = DWC_{5\times5}(O_{i}) \tag{12} \\
O_{i}^{3} = DWC_{7\times7}(O_{i}) \tag{13} \\
MSA_{i} = PWC(Channel\_Shuffle(O_{i}^{1}+O_{i}^{2}+O_{i}^{3})) \tag{14}
\end{gather}
where $i=1,2,3$.

\subsection{Feature Discovery Module}
To alleviate the omission risk of diminutive polyps, the Feature Discovery (FD) module~\cite{b50} aggregates and fuses all decoder features. As illustrated in Fig.~\ref{mynet}(d), through element-wise multiplication and addition operations, dependencies are established among the decoder features from different stages. First, the decoder features from stage 3, $CMD_3$, are upsampled using a 4×4 transpose convolution to match the size of the decoder features from stage 2, $CMD_2$. Then, element-wise multiplication and addition operations are applied to establish dependencies between $CMD_2$ and upsampled $CMD_3$, resulting in $G_{23}$. The same operation is also applied to $G_{23}$ and the decoder features from stage 1, $CMD_1$, to obtain the final output. This process can be expressed by the following formulas:
\begin{gather}
G_{23} = DEC_{4\times4}(CMD_{3})\times CMD_2+CMD_2 \tag{15} \\
output = DEC_{4\times4}(G_{23})\times CMD_1+CMD_1 \tag{16} 
\end{gather}
where $DEC_{4\times4}$ refers to a transposed convolution operation configured with a 4×4 kernel and padding size of 1.

\section{Experiments}

\subsection{Datasets}
Five widely recognized polyp segmentation datasets are employed: ClinicDB~\cite{b6}, EndoScene~\cite{b10}, ETIS~\cite{b8}, ColonDB~\cite{b9} and  Kvasir~\cite{b7}. The data partitioning strategy follows the same protocol as adopted in previous comparative studies. For training, a combined total of 1,450 images is selected, including 900 samples from Kvasir-SEG and 550 from CVC-ClinicDB. The unused images from these two datasets are allocated as visible test sets to examine the model’s learning capability. In contrast, the remaining three datasets function as unseen test sets to evaluate the network’s generalization ability across diverse clinical environments. To assess the effectiveness of CMFDNet, six quantitative indicators are adopted, including mIoU (mean IoU), $\mathrm{E}_\mathrm{\xi}$ (E-measure)~\cite{b31}, $\mathrm{F}_\mathrm{\beta}^\mathrm{w}$ (the weighted F-measure)~\cite{b29}, $\mathrm{S}_\mathrm{\alpha}$ (the structure measure)~\cite{b30}, MAE (mean absolute error), and mDic (mean Dice)~\cite{b28}.

\begin{table}[!t]
\centering
\setlength{\tabcolsep}{1.5pt}
\caption{CMFDNet's quantitative results, with the best scores shown in \textbf{bold}. The values are expressed in percentages(\%).}
\label{tab1}
\begin{tabular}{l|l|l|l|l|l|l|l|l} 
\hline
Dataset                   & Metric    & PraNet & \begin{tabular}[c]{@{}l@{}}FCB\\Former\end{tabular} & \begin{tabular}[c]{@{}l@{}}EC\\TransNet\end{tabular} & \begin{tabular}[c]{@{}l@{}}Swin\\Umanba\end{tabular} & \begin{tabular}[c]{@{}l@{}}VM-Unet\end{tabular} & \begin{tabular}[c]{@{}l@{}}VM-UnetV2\end{tabular} & Ours                  \\ 
\hline\hline
\multirow{6}{*}{ETIS}      & mDice       & 76.54  & 76.64                                               & 77.80                                                & 68.63                                                & 80.02                                             & 70.56                                              & \textbf{81.85 }          \\
                           & mIoU        & 68.38  & 68.64                                               & 69.67                                                & 60.84                                                & 72.36                                             & 62.82                                              & \textbf{\textbf{74.27}}  \\
                           & $F_\beta^w$ & 69.88  & 71.91                                               & 75.68                                                & 62.54                                                & 75.23                                             & 64.29                                              & \textbf{\textbf{77.53}}  \\
                           & $S_\alpha$  & 86.05  & 85.39                                               & 85.09                                                & 81.32                                                & 90.56                                             & 82.10                                              & \textbf{\textbf{94.12}}  \\
                           & $E_\xi$     & 87.99  & 89.88                                               & 88.64                                                & 83.25                                                & 90.31                                             & 84.94                                              & \textbf{\textbf{91.94}}  \\
                           & MAE         & 1.81   & 1.79                                                & 3.58                                                 & 2.52                                                 & 1.50                                              & 2.75                                               & \textbf{\textbf{1.35}}   \\ 
\hline
\multirow{6}{*}{Kvasir}    & mDice       & 90.81  & 91.57                                               & 90.83                                                & 90.72                                                & 91.42                                             & 90.79                                              & \textbf{91.74 }          \\
                           & mIoU        & 85.99  & 86.42                                               & 85.58                                                & 85.25                                                & 86.12                                             & 85.23                                              & \textbf{\textbf{87.15}}  \\
                           & $F_\beta^w$ & 89.17  & 90.34                                               & 89.49                                                & 88.97                                                & 89.67                                             & 89.39                                              & \textbf{\textbf{90.78}}  \\
                           & $S_\alpha$  & 91.91  & 92.04                                               & 91.73                                                & 91.63                                                & 91.87                                             & 91.89                                              & \textbf{\textbf{92.72}}  \\
                           & $E_\xi$     & 95.47  & 95.66                                               & 95.58                                                & 95.23                                                & 94.98                                             & 95.06                                              & \textbf{\textbf{96.14}}  \\
                           & MAE         & 2.68   & 2.46                                                & 89.39                                                & 3.02                                                 & 2.77                                              & 2.98                                               & \textbf{\textbf{2.43}}   \\ 
\hline
\multirow{6}{*}{ClinicDB}  & mDice       & 91.90  & 93.10                                               & 93.20                                                & 89.74                                                & 93.06                                             & 91.96                                              & \textbf{93.36 }          \\
                           & mIoU        & 87.04  & 88.12                                               & 88.50                                                & 84.75                                                & 88.46                                             & 87.23                                              & \textbf{\textbf{88.96}}  \\
                           & $F_\beta^w$ & 90.18  & 92.36                                               & 91.80                                                & 88.68                                                & 92.56                                             & 91.14                                              & \textbf{\textbf{92.62}}  \\
                           & $S_\alpha$  & 94.50  & 94.26                                               & 94.89                                                & 93.02                                                & 94.38                                             & 94.35                                              & \textbf{\textbf{95.54}}  \\
                           & $E_\xi$     & 97.42  & 97.78                                               & 97.88                                                & 95.41                                                & 96.49                                             & 97.25                                              & \textbf{\textbf{97.96}}  \\
                           & MAE         & 0.90   & 0.75                                                & 0.77                                                 & 1.72                                                 & 0.87                                              & 0.84                                               & \textbf{\textbf{0.74}}   \\ 
\hline
\multirow{6}{*}{ColonDB}   & mDice       & 76.16  & 80.73                                               & 77.80                                                & 76.92                                                & 81.50                                             & 79.86                                              & \textbf{83.05}           \\
                           & mIoU        & 69.07  & 72.61                                               & 69.67                                                & 68.61                                                & 73.23                                             & 72.00                                              & \textbf{\textbf{75.56}}  \\
                           & $F_\beta^w$ & 74.13  & 77.70                                               & 75.68                                                & 73.32                                                & 79.21                                             & 76.89                                              & \textbf{\textbf{81.44}}  \\
                           & $S_\alpha$  & 84.85  & 86.22                                               & 85.09                                                & 84.38                                                & 86.44                                             & 86.44                                              & \textbf{\textbf{88.01}}  \\
                           & $E_\xi$     & 87.68  & 91.65                                               & 88.64                                                & 89.00                                                & 90.91                                             & 90.29                                              & \textbf{\textbf{92.16}}  \\
                           & MAE         & 3.90   & 3.12                                                & 3.58                                                 & 3.87                                                 & 3.12                                              & 3.13                                               & \textbf{\textbf{2.99}}   \\ 
\hline
\multirow{6}{*}{EndoScene} & mDice       & 88.55  & 88.99                                               & 88.86                                                & 84.54                                                & 89.95                                             & 85.81                                              & \textbf{90.62 }          \\
                           & mIoU        & 81.32  & 81.91                                               & 82.38                                                & 72.56                                                & 83.88                                             & 77.75                                              & \textbf{\textbf{84.30}}  \\
                           & $F_\beta^w$ & 85.21  & 86.24                                               & 86.60                                                & 79.41                                                & 87.12                                             & 81.33                                              & \textbf{\textbf{88.36}}  \\
                           & $S_\alpha$  & 92.94  & 92.80                                               & 93.01                                                & 89.76                                                & 93.12                                             & 91.28                                              & \textbf{\textbf{94.12}}  \\
                           & $E_\xi$     & 96.02  & 96.31                                               & 97.05                                                & 93.49                                                & 96.40                                             & 94.04                                              & \textbf{\textbf{97.17}}  \\
                           & MAE         & 0.83   & 0.90                                                & 0.69                                                 & 1.19                                                 & 0.65                                              & 1.28                                               & \textbf{\textbf{0.58}}   \\
\hline
\end{tabular}
\end{table}

\subsection{Implementation Details}
CMFDNet is developed based on the PyTorch platform and executed on an NVIDIA A800 GPU for both training and inference. To ensure input uniformity, all images are standardized to a resolution of 224×224 pixels. The training phase lasts for 150 epochs with a mini-batch size set to 8. The optimization process utilizes the AdamW algorithm, beginning with an initial learning rate of $1 \times 10^{-4}$, which is reduced by half every 50 epochs as part of a scheduled decay strategy. To improve model robustness and generalization performance, diverse data augmentation techniques are employed, such as random horizontal flipping, rotational transformations, and color perturbations, thereby enriching the variability within the training data.

\subsection{Results}
\subsubsection{Quantitative Results.}
Six representative state-of-the-art models are selected for comparison, namely: VM-UnetV2~\cite{b42}, VM-Unet~\cite{b41}, Swin-UMamba~\cite{b40}, ECTransNet~\cite{b14}, FCBFormer~\cite{b39} and PraNet~\cite{b11}. For consistent evaluation, all competing methods were re-executed within the same experimental configuration. A summary of quantitative results is presented in Table~\ref{tab1}. When tested on the Kvasir and ClinicDB datasets—which are used to assess learning capability—CMFDNet achieves mDice scores that exceed the best-performing method by 0.17\% and 0.16\%, respectively, reflecting marginal gains in feature learning effectiveness. In contrast, on the ETIS, ColonDB, and EndoScene datasets, which emphasize generalization to unseen data, CMFDNet surpasses the highest SOTA benchmark by 1.83\%, 1.55\%, and 0.67\% in mDice score, respectively. These results demonstrate the superior generalization capacity of the proposed model.
Overall, CMFDNet consistently delivers better performance than the compared approaches in terms of both learning adaptability and cross-dataset generalization.

\subsubsection{Qualitative Results.}
\begin{figure}[!h]
\centering
\includegraphics[width=1\linewidth]{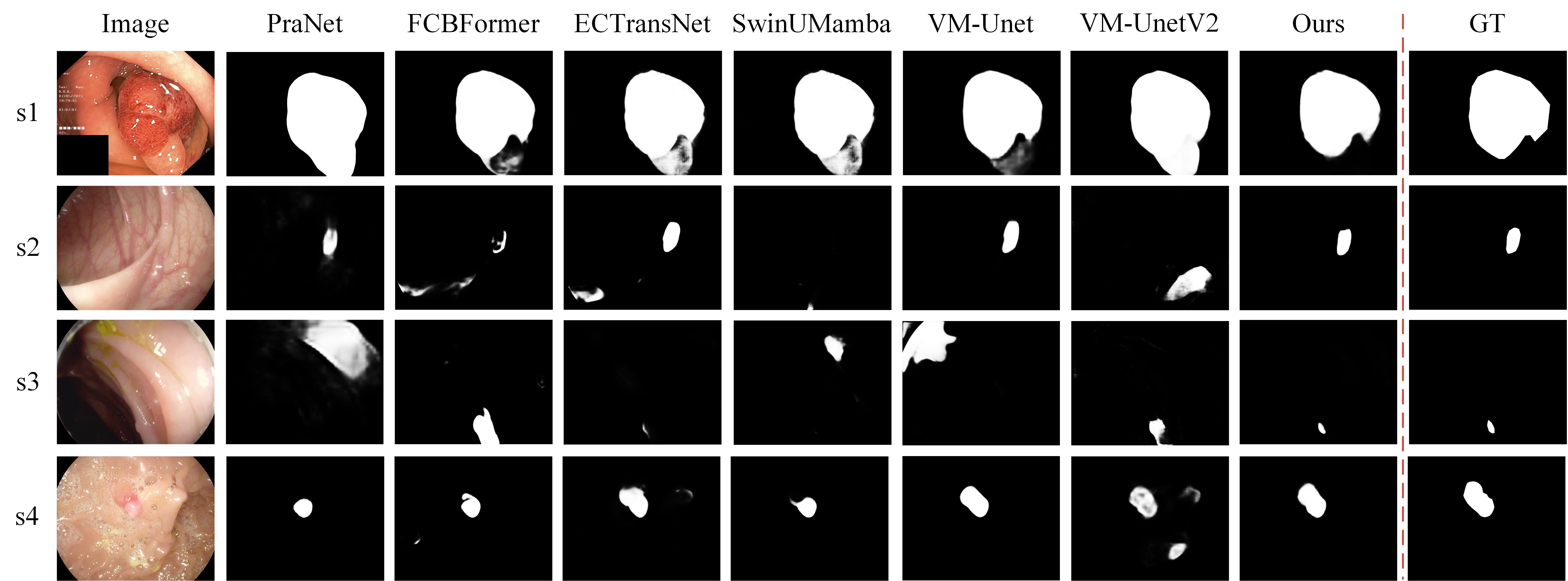}
\caption{The qualitative results of CMFDNet and SOTA methods.}
\label{compare}
\end{figure}

Fig.~\ref{compare} presents a qualitative comparison of segmentation outcomes produced by CMFDNet and several competitive baseline methods. Visually, CMFDNet demonstrates markedly enhanced delineation performance across all test cases. In cases s1, s2, and s3, CMFDNet delivers more precise delineation of polyps exhibiting diverse morphological characteristics and scale variations. Particularly in s3, where diminutive lesions are present, the model effectively identifies and segments small targets that are inadequately captured by other methods, thereby alleviating small-object omission. In s4, despite indistinct boundaries between the lesion and surrounding tissues, CMFDNet exhibits superior localization and structural discrimination, enabling more accurate region separation. Overall, these visual results confirm that CMFDNet achieves improved segmentation fidelity and boundary precision relative to these SOTA approaches.

\begin{table}[!h]
\centering
\setlength{\tabcolsep}{5pt}
\caption{The ablation experiment results of CMFDNet on ColonDB and ETIS. “Attention” refers to the attention blocks employed in the CMD and MSA modules.}\label{tab3}
\begin{tabular}{l|l|l|l|l|l|l|l|l|l} 
\hline
\multirow{2}{*}{CMD} & \multirow{2}{*}{MSA} & \multirow{2}{*}{FD} & \multirow{2}{*}{Attention} & \multicolumn{3}{l|}{ColonDB}                                               & \multicolumn{3}{l}{ETIS}                                                    \\ 
\cline{5-10}
                     &                      &                     &                      & mDice                   & mIoU                    & MAE                    & mDice                   & mIoU                    & MAE                     \\ 
\hline\hline
\XSolidBrush                     & \Checkmark                    & \Checkmark                      & GAB                  & 77.12                   & 70.31                   & 3.67                   & 76.23                   & 69.23                   & 1.85                    \\ 
\hline
\Checkmark                      & \XSolidBrush                     & \Checkmark                     & GAB                  & 81.94                   & 74.77                   & 3.12                   & 80.68                   & 73.10                   & 1.52                    \\ 
\hline
\Checkmark                    & \Checkmark                    & \XSolidBrush                     & GAB                  & 82.50                   & 74.67                   & 3.14                   & 81.05                   & 74.04                   & 1.47                    \\ 
\hline
\Checkmark                    & \Checkmark                    & \Checkmark                     & CBAM                 & 82.46                   & 74.89                   & 3.05                   & 81.21                   & 73.97                   & 1.41                    \\ 
\hline
\Checkmark                    & \Checkmark                    & \Checkmark                     & GAB                  & \textbf{\textbf{83.05}} & \textbf{\textbf{75.56}} & \textbf{\textbf{2.99}} & \textbf{\textbf{81.85}} & \textbf{\textbf{74.27}} & \textbf{\textbf{1.35}}  \\
\hline
\end{tabular}
\end{table}

\subsection{Ablation Experiments}

To assess the individual contributions of key components within CMFDNet, a series of ablation studies were performed on the ColonDB and ETIS datasets. Each module was selectively excluded to examine its functional impact, with the corresponding outcomes summarized in Table~\ref{tab3}. The results indicate that the absence of any single component negatively affects the overall segmentation performance. On ColonDB, excluding the CMD module led to the most substantial degradation, with mDice and mIoU dropping by 5.93\% and 5.25\%, respectively. This highlights the critical role of CMD, which serves as the central decoding structure of the network. Eliminating the MSA module resulted in moderate performance declines (–1.11\% mDice, –0.79\% mIoU), while omitting the FD module produced slightly smaller reductions (–0.55\% mDice, –0.89\% mIoU). Furthermore, when the GAB module was substituted with CBAM~\cite{b48}, the model’s accuracy decreased by 0.59\% in mDice and 0.67\% in mIoU, confirming the relative advantage of GAB in this architecture. Collectively, these findings underscore the necessity of each core component in maintaining CMFDNet’s high-level performance.

\section{Conclusions}
This paper introduces CMFDNet, a Cross-Mamba and Feature Discovery Network for Polyp Segmentation, constructed to mitigate key obstacles inherent in polyp segmentation tasks such as varying sizes and shapes, blurred boundaries, and the loss of small polyps. Experimental results demonstrate that CMFDNet achieves superior accuracy in polyp segmentation, which can reduce misdiagnosis and missed diagnosis of colorectal polyps during colonoscopy, thus enabling earlier clinical intervention and improving patient care efficiency. Despite the effectiveness of CMFDNet, it currently does not include a dedicated module for filtering noise, which is particularly important given the frequent presence of noise in polyp images. Enhancing robustness through explicit noise suppression remains a promising direction for future work.


\begin{thebibliography}{10}
\providecommand{\url}[1]{#1}
\csname url@samestyle\endcsname
\providecommand{\newblock}{\relax}
\providecommand{\bibinfo}[2]{#2}
\providecommand{\BIBentrySTDinterwordspacing}{\spaceskip=0pt\relax}
\providecommand{\BIBentryALTinterwordstretchfactor}{4}
\providecommand{\BIBentryALTinterwordspacing}{\spaceskip=\fontdimen2\font plus
\BIBentryALTinterwordstretchfactor\fontdimen3\font minus \fontdimen4\font\relax}
\providecommand{\BIBforeignlanguage}[2]{{%
\expandafter\ifx\csname l@#1\endcsname\relax
\typeout{** WARNING: IEEEtran.bst: No hyphenation pattern has been}%
\typeout{** loaded for the language `#1'. Using the pattern for}%
\typeout{** the default language instead.}%
\else
\language=\csname l@#1\endcsname
\fi
#2}}
\providecommand{\BIBdecl}{\relax}
\BIBdecl

\bibitem{b1}
M.~Sandler, A.~Howard, M.~Zhu, A.~Zhmoginov, and L.-C. Chen, ``Mobilenetv2: Inverted residuals and linear bottlenecks,'' in \emph{Proceedings of the IEEE conference on computer vision and pattern recognition}, 2018, pp. 4510--4520.

\bibitem{b6}
J.~Bernal, F.~J. S{\'a}nchez, G.~Fern{\'a}ndez-Esparrach, D.~Gil, C.~Rodr{\'\i}guez, and F.~Vilari{\~n}o, ``Wm-dova maps for accurate polyp highlighting in colonoscopy: Validation vs. saliency maps from physicians,'' \emph{Computerized medical imaging and graphics}, vol.~43, pp. 99--111, 2015.

\bibitem{b7}
D.~Jha, P.~H. Smedsrud, M.~A. Riegler, P.~Halvorsen, T.~De~Lange, D.~Johansen, and H.~D. Johansen, ``Kvasir-seg: A segmented polyp dataset,'' in \emph{MultiMedia modeling: 26th international conference, MMM 2020, Daejeon, South Korea, January 5--8, 2020, proceedings, part II 26}.\hskip 1em plus 0.5em minus 0.4em\relax Springer, 2020, pp. 451--462.

\bibitem{b8}
J.~Silva, A.~Histace, O.~Romain, X.~Dray, and B.~Granado, ``Toward embedded detection of polyps in wce images for early diagnosis of colorectal cancer,'' \emph{International journal of computer assisted radiology and surgery}, vol.~9, pp. 283--293, 2014.

\bibitem{b9}
N.~Tajbakhsh, S.~R. Gurudu, and J.~Liang, ``Automated polyp detection in colonoscopy videos using shape and context information,'' \emph{IEEE transactions on medical imaging}, vol.~35, no.~2, pp. 630--644, 2015.

\bibitem{b10}
D.~V{\'a}zquez, J.~Bernal, F.~J. S{\'a}nchez, G.~Fern{\'a}ndez-Esparrach, A.~M. L{\'o}pez, A.~Romero, M.~Drozdzal, and A.~Courville, ``A benchmark for endoluminal scene segmentation of colonoscopy images,'' \emph{Journal of healthcare engineering}, vol. 2017, no.~1, p. 4037190, 2017.

\bibitem{b11}
D.-P. Fan, G.-P. Ji, T.~Zhou, G.~Chen, H.~Fu, J.~Shen, and L.~Shao, ``Pranet: Parallel reverse attention network for polyp segmentation,'' in \emph{International conference on medical image computing and computer-assisted intervention}.\hskip 1em plus 0.5em minus 0.4em\relax Springer, 2020, pp. 263--273.

\bibitem{b14}
W.~Liu, Z.~Li, C.~Li, and H.~Gao, ``Ectransnet: An automatic polyp segmentation network based on multi-scale edge complementary,'' \emph{Journal of Digital Imaging}, vol.~36, no.~6, pp. 2427--2440, 2023.

\bibitem{b16}
W.~Wang, H.~Sun, and X.~Wang, ``From coarse to fine: A novel colon polyp segmentation method like human observation,'' in \emph{Chinese Conference on Pattern Recognition and Computer Vision (PRCV)}.\hskip 1em plus 0.5em minus 0.4em\relax Springer, 2024, pp. 264--278.

\bibitem{b17}
H.~Sung, J.~Ferlay, R.~L. Siegel, M.~Laversanne, I.~Soerjomataram, A.~Jemal, and F.~Bray, ``Global cancer statistics 2020: Globocan estimates of incidence and mortality worldwide for 36 cancers in 185 countries,'' \emph{CA: A Cancer Journal for Clinicians}, vol.~71, no.~3, pp. 209--249, 2021.

\bibitem{b20}
J.~Long, E.~Shelhamer, and T.~Darrell, ``Fully convolutional networks for semantic segmentation,'' in \emph{Proceedings of the IEEE Conference on Computer Vision and Pattern Recognition (CVPR)}, June 2015.

\bibitem{b21}
O.~Ronneberger, P.~Fischer, and T.~Brox, ``U-net: Convolutional networks for biomedical image segmentation,'' in \emph{Medical image computing and computer-assisted intervention--MICCAI 2015: 18th international conference, Munich, Germany, October 5-9, 2015, proceedings, part III 18}.\hskip 1em plus 0.5em minus 0.4em\relax Springer, 2015, pp. 234--241.

\bibitem{b28}
F.~Milletari, N.~Navab, and S.-A. Ahmadi, ``V-net: Fully convolutional neural networks for volumetric medical image segmentation,'' in \emph{2016 Fourth International Conference on 3D Vision (3DV)}, 2016, pp. 565--571.

\bibitem{b29}
R.~Margolin, L.~Zelnik-Manor, and A.~Tal, ``How to evaluate foreground maps?'' in \emph{Proceedings of the IEEE Conference on Computer Vision and Pattern Recognition (CVPR)}, June 2014.

\bibitem{b30}
D.-P. Fan, M.-M. Cheng, Y.~Liu, T.~Li, and A.~Borji, ``Structure-measure: A new way to evaluate foreground maps,'' in \emph{Proceedings of the IEEE International Conference on Computer Vision (ICCV)}, Oct 2017.

\bibitem{b31}
D.-P. Fan, C.~Gong, Y.~Cao, B.~Ren, M.-M. Cheng, and A.~Borji, ``Enhanced-alignment measure for binary foreground map evaluation,'' \emph{arXiv preprint arXiv:1805.10421}, 2018.

\bibitem{b27}
{Wang, Wei and Sun, Huiying and Wang, Xin}, ``Lssnet: A method for colon polyp segmentation based on local feature supplementation and shallow feature supplementation,'' in \emph{International Conference on Medical Image Computing and Computer-Assisted Intervention}.\hskip 1em plus 0.5em minus 0.4em\relax Springer, 2024, pp. 446--456.

\bibitem{b32}
P.~Screening and P.~E. Board, ``Colorectal cancer screening (pdq®): Health professional version,'' \emph{PDQ Cancer Information Summaries [Internet]}, Oct 2024.

\bibitem{b34}
A.~Dosovitskiy, L.~Beyer, A.~Kolesnikov, D.~Weissenborn, X.~Zhai, T.~Unterthiner, M.~Dehghani, M.~Minderer, G.~Heigold, S.~Gelly \emph{et~al.}, ``An image is worth 16x16 words: Transformers for image recognition at scale,'' \emph{arXiv preprint arXiv:2010.11929}, 2020.

\bibitem{b35}
X.~Liu, C.~Zhang, and L.~Zhang, ``Vision mamba: A comprehensive survey and taxonomy,'' \emph{arXiv preprint arXiv:2405.04404}, 2024.

\bibitem{b36}
Y.~Liu, Y.~Tian, Y.~Zhao, H.~Yu, L.~Xie, Y.~Wang, Q.~Ye, J.~Jiao, and Y.~Liu, ``Vmamba: Visual state space model,'' \emph{Advances in neural information processing systems}, vol.~37, pp. 103\,031--103\,063, 2024.

\bibitem{b38}
Z.~Qiu, Z.~Wang, M.~Zhang, Z.~Xu, J.~Fan, and L.~Xu, ``Bdg-net: boundary distribution guided network for accurate polyp segmentation,'' in \emph{Medical Imaging 2022: Image Processing}, vol. 12032.\hskip 1em plus 0.5em minus 0.4em\relax SPIE, 2022, pp. 792--799.

\bibitem{b39}
E.~Sanderson and B.~J. Matuszewski, ``Fcn-transformer feature fusion for polyp segmentation,'' in \emph{Annual conference on medical image understanding and analysis}.\hskip 1em plus 0.5em minus 0.4em\relax Springer, 2022, pp. 892--907.

\bibitem{b40}
J.~Liu, H.~Yang, H.-Y. Zhou, Y.~Xi, L.~Yu, C.~Li, Y.~Liang, G.~Shi, Y.~Yu, S.~Zhang \emph{et~al.}, ``Swin-umamba: Mamba-based unet with imagenet-based pretraining,'' in \emph{International Conference on Medical Image Computing and Computer-Assisted Intervention}.\hskip 1em plus 0.5em minus 0.4em\relax Springer, 2024, pp. 615--625.

\bibitem{b41}
J.~Ruan, J.~Li, and S.~Xiang, ``Vm-unet: Vision mamba unet for medical image segmentation,'' \emph{arXiv preprint arXiv:2402.02491}, 2024.

\bibitem{b42}
M.~Zhang, Y.~Yu, S.~Jin, L.~Gu, T.~Ling, and X.~Tao, ``Vm-unet-v2: rethinking vision mamba unet for medical image segmentation,'' in \emph{International Symposium on Bioinformatics Research and Applications}.\hskip 1em plus 0.5em minus 0.4em\relax Springer, 2024, pp. 335--346.

\bibitem{b43}
S.~K. Bhat and J.~E. East, ``Colorectal cancer: prevention and early diagnosis,'' \emph{Medicine}, vol.~43, no.~6, pp. 295--298, 2015.

\bibitem{b45}
Z.~Wu, X.~Xiong, Y.~Chen, S.~Li, and H.~Chen, ``Moe-polyp: Shifting more attention to small polyp segmentation via mixture-of-experts,'' in \emph{Proceedings of the 6th ACM International Conference on Multimedia in Asia}, 2024, pp. 1--1.

\bibitem{b46}
S.~Zhao, H.~Chen, X.~Zhang, P.~Xiao, L.~Bai, and W.~Ouyang, ``Rs-mamba for large remote sensing image dense prediction,'' \emph{IEEE Transactions on Geoscience and Remote Sensing}, 2024.

\bibitem{b47}
X.~Zhang, X.~Zhou, M.~Lin, and J.~Sun, ``Shufflenet: An extremely efficient convolutional neural network for mobile devices,'' in \emph{Proceedings of the IEEE conference on computer vision and pattern recognition}, 2018, pp. 6848--6856.

\bibitem{b48}
S.~Woo, J.~Park, J.-Y. Lee, and I.~S. Kweon, ``Cbam: Convolutional block attention module,'' in \emph{Proceedings of the European conference on computer vision (ECCV)}, 2018, pp. 3--19.

\bibitem{b49}
M.~M. Rahman, M.~Munir, and R.~Marculescu, ``Emcad: Efficient multi-scale convolutional attention decoding for medical image segmentation,'' in \emph{Proceedings of the IEEE/CVF Conference on Computer Vision and Pattern Recognition}, 2024, pp. 11\,769--11\,779.

\bibitem{b50}
B.~Xiao, J.~Hu, W.~Li, C.-M. Pun, and X.~Bi, ``Ctnet: Contrastive transformer network for polyp segmentation,'' \emph{IEEE Transactions on Cybernetics}, vol.~54, no.~9, pp. 5040--5053, 2024.

\end{thebibliography}

\end{document}